\documentclass[sigconf]{acmart}
\usepackage{booktabs} 
\usepackage{xspace}

\usepackage{subfig}

\usepackage{multirow} 
\usepackage{array} 
\usepackage{tabularx} 
\usepackage{enumitem}

\graphicspath{{Figures/}}

\newcommand{\ie}{\emph{i.e.,}\xspace}
\newcommand{\eg}{\emph{e.g.,}\xspace}

\newcommand{\paratitle}[1]{\vspace{1ex}\noindent \textbf{#1}}

\setcopyright{rightsretained}

\acmDOI{}

\acmISBN{}

\acmConference[]{Conference}{2019}{}
\acmYear{}
\copyrightyear{}

\acmPrice{15.00}

\begin{document}

\title{Understanding the Stability of Medical Concept Embeddings}

\author{Grace E. Lee}
\orcid{}
\affiliation{%
  \institution{School of Computer Science and Engineering\linebreak Nanyang Technological University, Singapore}
  \streetaddress{}
  \city{}
  \state{}
  \postcode{}
}
\email{leee0020@e.ntu.edu.sg}

\author{Aixin Sun}
\orcid{}
\affiliation{%
  \institution{School of Computer Science and Engineering\linebreak Nanyang Technological University, Singapore}
  \streetaddress{}
  \city{}
  \state{}
  \postcode{}
}
\email{axsun@ntu.edu.sg}

\begin{abstract}
Frequency is one of the major factors for training quality word embeddings. Several work has recently discussed the stability of word embeddings in general domain and suggested factors influencing the stability. In this work, we conduct a detailed analysis on the stability of concept embeddings in medical domain, particularly the relation with concept frequency. The analysis reveals the surprising high stability of low-frequency concepts: low-frequency (<100) concepts have the same high stability as high-frequency (>1000) concepts. To develop a deeper understanding of this finding, we propose a new factor, \textit{the noisiness of context words}, which influences the stability of medical concept embeddings, regardless of frequency. We evaluate the proposed factor by showing the linear correlation with the stability of medical concept embeddings. The correlations are clear and consistent with various groups of medical concepts. Based on the linear relations, we make suggestions on ways to adjust the noisiness of context words for the improvement of stability. Finally, we demonstrate that the proposed factor extends to the word embedding stability in general domain.
\end{abstract}

\keywords{Medical concept embedding, Stability, Context words, Frequency,  UMLS}

\maketitle

\section{Introduction}
\label{sec:Introduction}

Medical concepts are medical terminologies linked to Unified Medical Language System (UMLS)\footnote{UMLS is the metathesaurus in medical domain which provides an unified language system over various medical dictionaries such as ICD-10, MeSH, and SNOMED-CT. More details are in \url{https://www.nlm.nih.gov/research/umls/}}. 
They are frequently used in medical domain for accurate and effective communication.
Using medical concepts identified by unique IDs (\eg CUI in UMLS) helps avoid misunderstanding and allows to utilize semantic information in UMLS.

Word embeddings encode syntactic and semantic aspects of words into low-dimensional and dense representations. As they have shown great advance, many studies have proposed approaches to learning medical concept embeddings using various biomedical resources.

Several studies have shown the instability of word embeddings in general domain ~\cite{Instability_downstream,NAACL18_stability,TACL18_stability,COLING_16_instable_optimal_parameters}.
Whenever word embeddings are trained on the same training corpus with the same parameter setting, encoded relations among words should be consistent in each time.
It is reported however that word embeddings fail to learn consistent relations, shown by using a stability measure (more details later).

The instability of embeddings poses greater problems in medical domain.
Unstable medical concept embeddings result in a wide-range of negative consequences from unsuccessful document filtering to misdiagnosis. 
While medical concept embeddings are of importance, their stability has not been carefully investigated.
In this work, we focus on medical concept embeddings and examine their stability.  
We use `word embedding' and `(medical) concept embedding' interchangeably throughout this work.

It is a well-known premise that frequency is one of the major factors contributing to quality word embeddings. 
As word embeddings are trained using word co-occurrence information, 
a high word frequency in a corpus provides sufficient training instances and thus helps obtain a quality word embedding. 
We conduct a detailed analysis on the relation between the stability of medical concept embeddings and concept frequency.
The analysis reveals that some low-frequency concepts result in high stability as high-frequency concepts do. 
For example, the frequencies of two concepts, \textsf{Anaesthesia procedure} and \textsf{Knee joint operation} are 10,377 and 32 respectively in our dataset and both concepts have the same high level of stability 0.8.
Our analysis results show that there are other factors that influence the stability of medical concept embeddings in addition to frequency.

Finally, we propose a new factor, \textit{the noisiness of context words}, which affects the stability of medical concept embeddings. 
The rationale is: if the context words of concept are noisy, its embedding undergoes inconsistent training in various directions, leading to the low quality of embedding. 
We use a measure called \textit{normalized entropy} to compute the noisiness of context words. 
It quantifies how flat or peaked a distribution of elements (context words in this context) is.
When a word has a high frequency in a corpus, its number of context words is also high.
Normalized entropy negates the effect of the total number of context words using normalization.
In this way, we measure solely the noisiness of context words without taking frequency into account.

In medical domain word2vec is widely adopted when training medical concepts embeddings. 
We focus on word2vec~\cite{word2vec_original} as a representative among existing word embedding algorithms. 
In the experiments, we demonstrate the relation between the proposed factor and the stability of concept embeddings by calculating a linear correlation coefficient.
The evaluation result shows moderate correlations. 
The correlations are clear and consistent with various groups of medical concepts. 
The proposed factor provides an empirical reasoning of how low-frequency concepts are able to result in high-stability embeddings, despite the small number of training instances.  
Besides, based on the moderate linear correlation, we make suggestions on ways to adjust the noisiness of context words. 
Lastly, we extend the evaluation to general-domain word embeddings and show the equivalent linear relation of the proposed factor.

\section{Preliminaries and Related Work}
\label{sec:Related Work}

In this section, we first review word embedding algorithms, and then introduce  training medical concept embeddings with different types of medical data. Next, we provide an overview of previous studies on the stability of word embeddings.

\subsection{Word Embeddings}
Word embeddings are a fundamental building block in NLP tasks. 
It represents syntactic and semantic meanings of a word into a low-dimensional vector of real values.
There are several word embedding algorithms such as word2vec~\cite{word2vec_original}, GloVe~\cite{GloVe_original}, and FastText~\cite{FastText_subword_info}. 
While details in training word embeddings are different in algorithms, they share the same hypothesis: words that occur in the same contexts tend to have similar meanings.

In this paper, we focus on word2vec, particularly the skip-gram model with negative sampling~\cite{word2vec_original}, as it is a popular model to train embeddings in medical domain. 
It trains an embedding by using co-occurrences between a given word and its neighboring words (context words) that appear within a pre-defined window size.
The objective of skip-gram model is to predict the context words given target word. 
Initially, word embeddings are randomly initialized. During the training, the skip-gram model find a set of word embeddings, which maximizes the objective function: 
\begin{equation}
\label{eq:skip_gram_objective}
\frac {1} {T}\sum _{i=1}^{T}\sum _{-k\leq j\leq k,j\neq 0}\log P\left( w_{i+j}|w_{i}\right)
\end{equation}
where $k$ is a window size, and $w_{i+j}$ indicates a context word for a given target word $w_{i}$ in distance of $j$. The sign of $j$ indicates a preceding or following direction of $w_{i}$.
Whenever word embeddings are trained, the absolute values in word embeddings are different due to the random initialization of word embeddings by design. In other words, they are in different embedding spaces. However, they are supposed to encode the consistent meanings of words because they are trained on the same training corpus with the same parameter setting.

Contextual word embeddings (\eg BERT, ELMo) have advanced word representations in recent years. 
They necessitate computational power and large amounts of data to train and/or fine-tune thousands of parameters. 
In medical domain, these conditions are challenging to meet because a large scale of medical data (\eg clinical records) is often restricted to the public over privacy concerns.
Thus, word embeddings are still frequently used tools to start off with. 
They are computationally inexpensive and fast, leading to faster prototyping and development.
Moreover, word embeddings show effective performance even when a training corpus is small.
Previous literature reports that using a small and topic-specific corpus is more effective than using a large and general corpus for domain-specific word embeddings~\cite{ACL17_workshop_corpus_specificity_negative_result,CliEmbed_eval_Howto_train_embeddings_biomedical,CliEmbed_eval_LargeCorpus_not_enhance}.

\subsection{Medical Concept Embeddings}

Medical concepts (\ie medical terminologies linked to a knowledge base UMLS) are so ubiquitous that they are in diverse medical applications/resources, such as electronic health records, health insurance claims, and biomedical literature. 
Each medical resource provides additional information specialized in its own application, in addition to medical concepts.
Previous studies have developed various approaches for effective learning of concept embeddings by leveraging extra information given by different resources.

Electronic health records (EHRs) are patients' visit records to hospitals. Each record consists of medical concepts related to given visit, as well as patient's demographic information and medical history. Records are also tagged with timestamps of visits. 
EHRs enable effective learning of concept embeddings by incorporating information shared by simialr patients or temporal information, in addition to co-occurrences of concepts within records~\cite{CliConEmbed_first_KDD16_EHRs,CliConEmbed_second_KDD17_EHRs,CliConEmbed_ijcai18_EHSs}. 
Health insurance claims are similar to  EHRs and each claim includes medical concepts tagged with temporal information. 
However, the occurrences of claims tend to be sparse and irregular compared to EHRs. 
Sporadic claims make it challenging to learn the relatedness between medical concepts into embeddings. In~\cite{CliConEmbed_claim_data_clinical_narrative}, the authors have introduced grouping and shuffling techniques to mitigate the challenge.

Lastly, biomedical literature is in formal written language and there is a significant body of work to train concept embeddings on biomedical literature~\cite{CliConEmbed_improve_labeld_corpora,CliConEmbed_improve_joint_text_entity,CliConEmbed_ACL16workshop_retrofitting}.
Unlike the previous two resources, it does not have pre-tagged medical concepts in it. To train concept embeddings, concepts must be identified first by using off-the-shelf tools such as MetaMap~\cite{MetaMap}, QuickUMLS~\cite{quickUMLS}. 
The most simple and straightforward approach using biomedical literature is called (\textbf{cui2vec})~\cite{cui2vec_beam_et_al}. It considers each concept a single word and apply a word embedding algorithm to concepts and their surrounding non-medical words. 
Another similar approach (\textbf{NLM}) is to apply an algorithm to only concepts, eliminating non-medical words~\cite{CIKM14short_De_Vine}. Table~\ref{tab:sentence} presents example input texts of the two approaches. A word embedding algorithm is then applied to the input text to train concept embeddings.

\begin{table}
	\centering
	\caption{Examples of input text for training concept embeddings. Medical concepts are presented in square brackets. A word embedding algorithm such as word2vec is applied.}
	\label{tab:sentence}
	\begin{tabular}{p{0.52in}|p{2.50in}}
		\toprule
		\textbf{Original sentence }& Calcium carbonate appears to be as effective as aluminum hydroxide in binding dietary phosphorus in hemodialysis patients. \\
		\midrule
		\midrule
		\textbf{cui2vec}& \textsf{[calcium\_carbonate]} appears to be as effective as \textsf{[aluminum\_hydroxide]} in binding dietary \textsf{[phosphorus]} in \textsf{[hemodialysis]} patients.\\ 
		\midrule
		\textbf{NLM}& \textsf{[calcium\_carbonate]} \textsf{[aluminum\_hydroxide]} \textsf{[phosphorus]} \textsf{[hemodialysis]}\\
		\bottomrule
	\end{tabular}
\end{table}

Among a wide range of approaches for medical concept embeddings, in this work we use \textbf{cui2vec} and \textbf{NLM} to train medical concept embeddings. 
They are general enough to be adopted not only in biomedical literature but also in EHRs and insurance claims. 
For example, NLM can be directly applied for EHRs and insurance claims where each record/claim consists of a bag of medical concepts.
Lastly, while EHRs and health insurance claims are often privately owned, biomedical literature is publicly available.

\subsection{Instability of Word Embeddings}

In recent years, there have been several studies reporting the instability of word embeddings in general domain. 
It is shown that when word embeddings are trained with the same corpus and hyperparameters but with random weight initialization in multiple times, the relations among words, in particular the nearest neighbor words, are different across the sets of word embeddings. Thus, the word embeddings are considered unstable.

Existing studies have suggested several factors that influence the stability of word embeddings. They are classified into two groups: corpus-level factors and word-level factors.
As corpus-level factors, in~\cite{TACL18_stability}, the authors explore various values of a corpus size and different document lengths in the corpus. 
In~\cite{CliEmbed_eval_LargeCorpus_not_enhance,ACL17_workshop_corpus_specificity_negative_result,NAACL18_stability}, it is shown that a small and topic-specific corpus helps train more stable word embeddings than a large and general corpus.

As word-level factors, word frequency, part-of-speech (POS) tags, and concreteness of word's meaning are studied~\cite{NAACL18_stability,concreteness_noun}. In~\cite{NAACL18_stability} the authors discover that words in some POS tags tend to have more stable embeddings than other tags. They also report that frequency is not a major factor on the stability of word embeddings. However, the authors do not further examine the reasons behind the finding.

Across corpus-level and word-level factors, most factors are related to word frequency either directly or indirectly. 
For example, in corpus-level factors, different sizes of corpus and topic specificity in a corpus incur the changes of word frequency. 
In word-level factors, part-of-speech tags are also related to word frequency, since some part-of-speech tags appear more frequently than other tags in a corpus.

\section{Analysis}
\label{sec:Analysis_Stability_Frequency}

Frequency is one of the well-known factors that influence the quality of word embeddings.  Fundamentally word embedding algorithms (\eg word2vec and GloVe) train word embeddings using co-occurrence information of words. A high frequency of a  word in a corpus is equivelent to a large number of co-occurred words (\ie training instances), so that it helps produce a quality word embedding. In this section, we conduct a datailed analysis on the stability of medical concept embeddings with concept frequency in a training corpus. 
We first introduce the stability measure and describe the settings used in training medical concept embeddings. Then, we discuss the analysis.

\subsection{Stability Measure}
\label{s_sec:stability_meas}

We use the stability measure to evaluate the quality of medical concept embeddings~\cite{TACL18_stability,NAACL18_stability}. It is initially proposed for word embedding in general domain. We first explain the stability in the context of general-domain word embeddings. Next, we present how it is applied to medical concept embeddings.

A stability value represents how stable a word embedding is. 
It is the portion of overlapping words between the $n$ nearest neighbors from different embedding spaces. 
Each embedding space is obtained by training word embeddings with the same training corpus and hyperparameters but with random initializations of embeddings and random sampling in a negative sampling technique, if it is used in word2vec.

Formally, given a word $w$ and two embedding spaces, $P$ and $Q$, let  $P_w$  and $Q_w$ be the $n$ nearest neighbors of $w$ in $P$ and $Q$ based on embedding similarities, respectively.
The stability value for word $w$'s embedding is the ratio of overlapping words in $P_w$ and $Q_w$.
\begin{equation}
\label{eqn:stability}
stability(w)=\frac{|P_w \cap Q_w|}{n}
\end{equation}
If $P_w$ and $Q_w$ consist of the same set of words, $stability(w)=1.0$. It demonstrates the embedding of $w$ encodes consistent semantic information of $w$. Hence, $w$'s embedding is considered stable.

In our problem setting, we replace a word $w$ with a medical concept $c$. The stability of concept embeddings is defined as the portion of overlapping nearest concepts from the three spaces $P$, $Q$, and $R$, \ie $\frac{|P_c \cap Q_c \cap R_c|}{n}$, instead of two spaces. In doing so, the stability of medical concept embeddings is evaluated by stricter conditions, since often biomedical applications require a high level of accuracy. The number of the nearest neighbors is set to 10  (\(n\)=10), and  the similarity between concept embeddings is calculated by cosine similarity~\cite{NAACL18_stability}.

\subsection{Training Medical Concept Embeddings}
\label{s_sec:medical_concept_embed}

There have been many studies on approaches for training medical concept embeddings with various biomedical data  (details in Related Work). In this work, we examine two approaches, cui2vec~\cite{cui2vec_beam_et_al} and NLM~\cite{CIKM14short_De_Vine}, using biomedical literature data.

\paratitle{Dataset.}
We use OHSUMED dataset to train medical concept embeddings. It consists of 348,566 abstracts sampled from a MEDLINE corpus. Though a MEDLINE corpus is much larger, multiple studies report that compared to a large and general corpus, a small and topic-specific corpus is more effective for embeddings in domain-specific applications~\cite{CliEmbed_eval_Howto_train_embeddings_biomedical,CliEmbed_eval_LargeCorpus_not_enhance}. For medical concept identification, there are several tools available such as MetaMap~\cite{MetaMap}, NCBO BioPortal~\cite{NCBO_bioportal}, PubTator~\cite{PubTator}, and QuickUMLS~\cite{quickUMLS}. We use QuickUMLS because it is faster than other tools. Total 40,625 (unique) medical concepts are extracted.

\begin{figure}[t]
	\centering
	\includegraphics[width=.85\columnwidth]{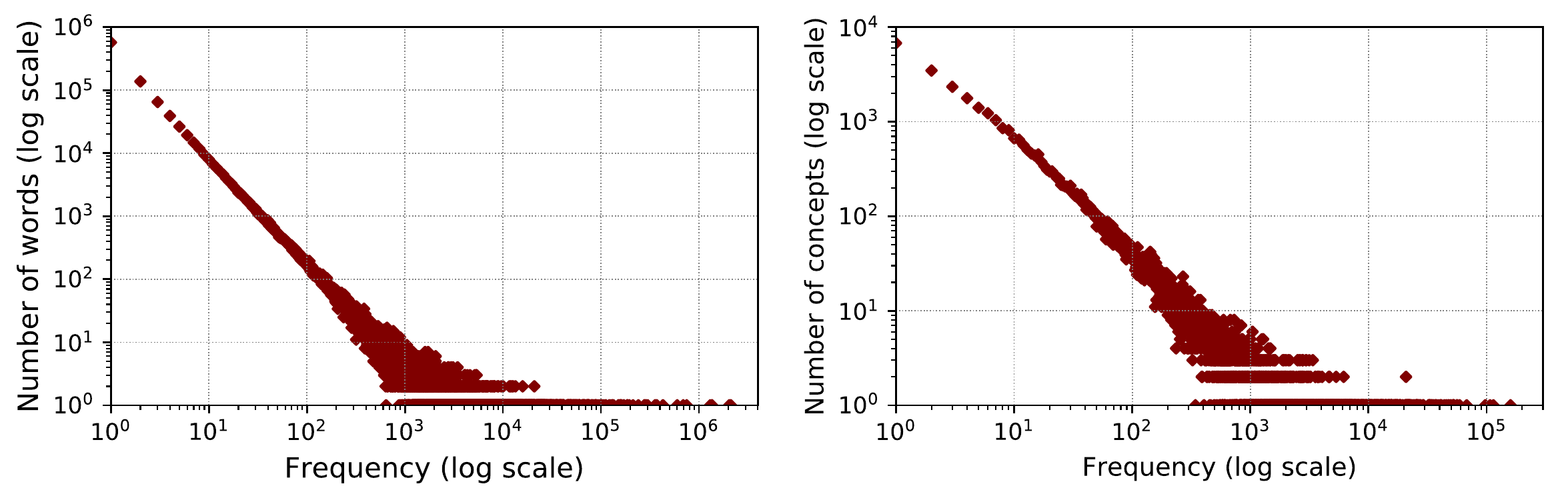}
	\caption{Distributions of words (left) and medical concepts (right) over frequencies in
		the OHSUMED dataset. The majority of concepts have low frequency.}
	\label{fig:wordconceptfreq}
\end{figure}

The vast majority of concepts have low frequencies. 
Figure~\ref{fig:wordconceptfreq} shows a comparison between the frequency distributions of words (left) and the extracted concepts (right). As expected, word frequencies follow a power law distribution. The distribution of concept frequencies also shows that the very similar distribution of words. 
When we consider that a high frequency helps train quality word embedding, most concepts are likely to have low-quality embeddings due to their low frequency.

Finally, the original sentences are transformed with medical concepts, as shown in Table~\ref{tab:sentence}. A word embedding algorithm is applied and trains concept embeddings.

\paratitle{Word2vec setting.}
We use the skip-gram model with negative sampling implemented in Gensim\footnote{\url{https://radimrehurek.com/gensim/index.html}}. The embedding size is set to 200 and the minimum frequency is set to 5. 
The window size for context words and  the number of epochs are set to 7 and 50, respectively. 
The remaining hyperparameters are set to default values. 
After rare words are excluded using the minimum frequency threshold, total 25,491 concept embeddings are trained.

Different hyperparameter values affect the quality of word embeddings~\cite{word2vec_original,Levy_w2v_parameter_TACL,RecSys_hyperParas}. 
We explore different values for five hyperparameters: the context window size, epochs, the number of negative samples, the smoothing parameter for the negative sample distribution, and the subsampling rate. We examine the changes of embedding stability when different values are used in the training. However, the evaluation results show that they result in  marginal changes and the average stabilities are very similar. We report the results in Appendix.

\subsection{Analysis: Stability of Medical Concept Embeddings and Frequency}
\label{sub-sec:Stab_Observation}
So far, we have introduced the stability measure and the definition of stability used in this work. We have also described the setting in training concept embeddings.
Now, we present a detailed analysis on the stability of concept embeddings with frequency and discuss their relations.

\begin{figure}[t]
	\centering
	\includegraphics[width=0.85\columnwidth]{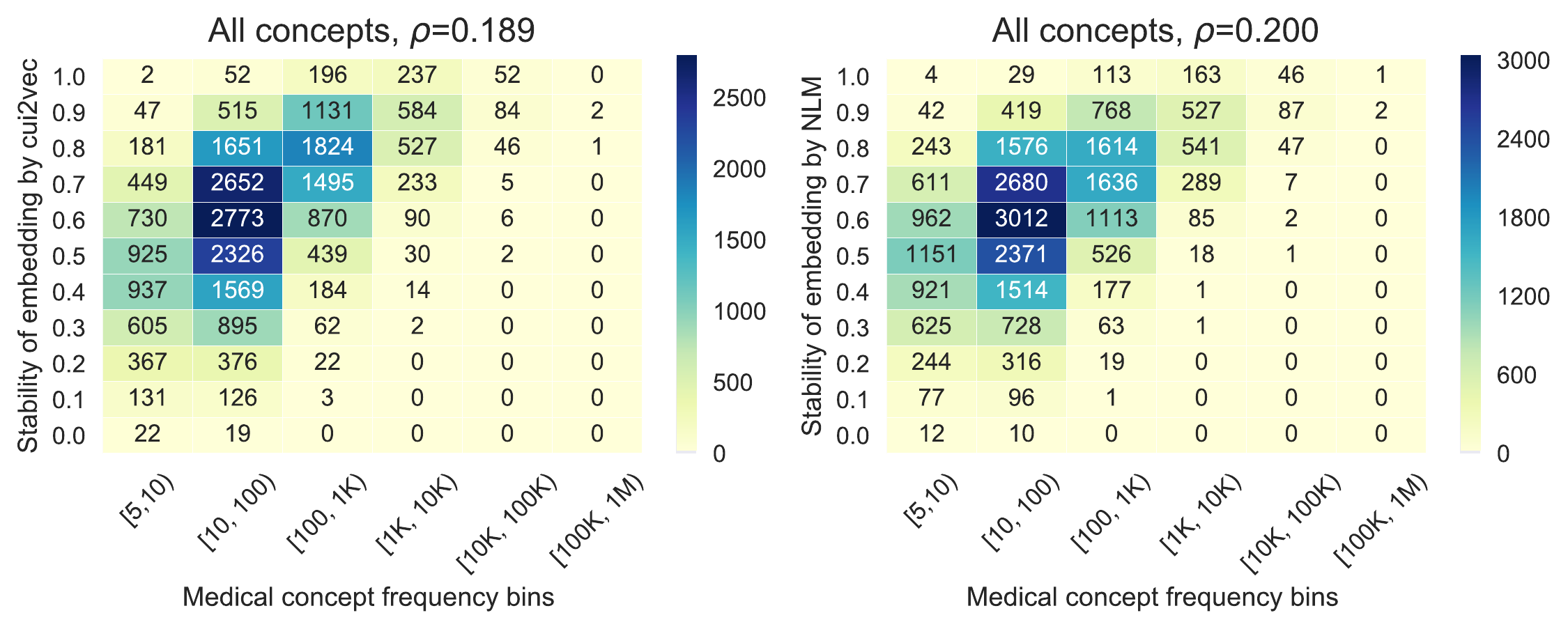}
	\caption{Distributions of medical concepts over frequency and the stability of embeddings trained by cui2vec and NLM in heatmaps. Total number of concepts is 25,491. Each cell indicates the number of medical concepts in the corresponding bin intersected by frequency and stability.}
	\label{fig:heatmaps_all_con}
\end{figure}

Figure~\ref{fig:heatmaps_all_con} shows a distribution of medical concept embeddings over frequency and embedding stability, trained by cui2vec and NLM.
In a subfigure, the $x$-axis indicates the ranges of concept frequency, and the $y$-axis indicates the stability from 0.0 to 1.0. Each cell denotes the number of medical concepts that belong to the corresponding frequency and stability bins. The color of cells shows the relative number of concepts.

As shown in Figure~\ref{fig:heatmaps_all_con}, high-frequency concepts tend to have high-stability concept embeddings. When a frequency is greater than 1,000, most concepts have stability higher than 0.5. The result confirms that indeed high frequency helps produce high stability embeddings with the sufficient training instances. Overall, cui2vec and NLM show similar distributional patterns.

We observe interesting stability distributions among low-frequency concepts.
The stability of low-frequency concepts (frequency $<100$) varies widely from 0.0 to 1.0. 
They end up with the high stability embeddings even with the much smaller number of training instances than high-frequency concepts. 
For example, here are two medical concepts, \textsf{Anaesthesia procedure} (CUI: C0002903) and \textsf{Knee joint operation} (CUI: C0187769), with drastically different frequencies. \textsf{Anaesthesia procedure} has top 10\% frequency (frequency: 10,377) and \textsf{Knee joint operation} has bottom 10\% frequency (frequency: 32) in the corpus. However, both result in the very high stability 0.8 in cui2vec and NLM.

\begin{figure}
	\centering
	\includegraphics[width=0.85\columnwidth]{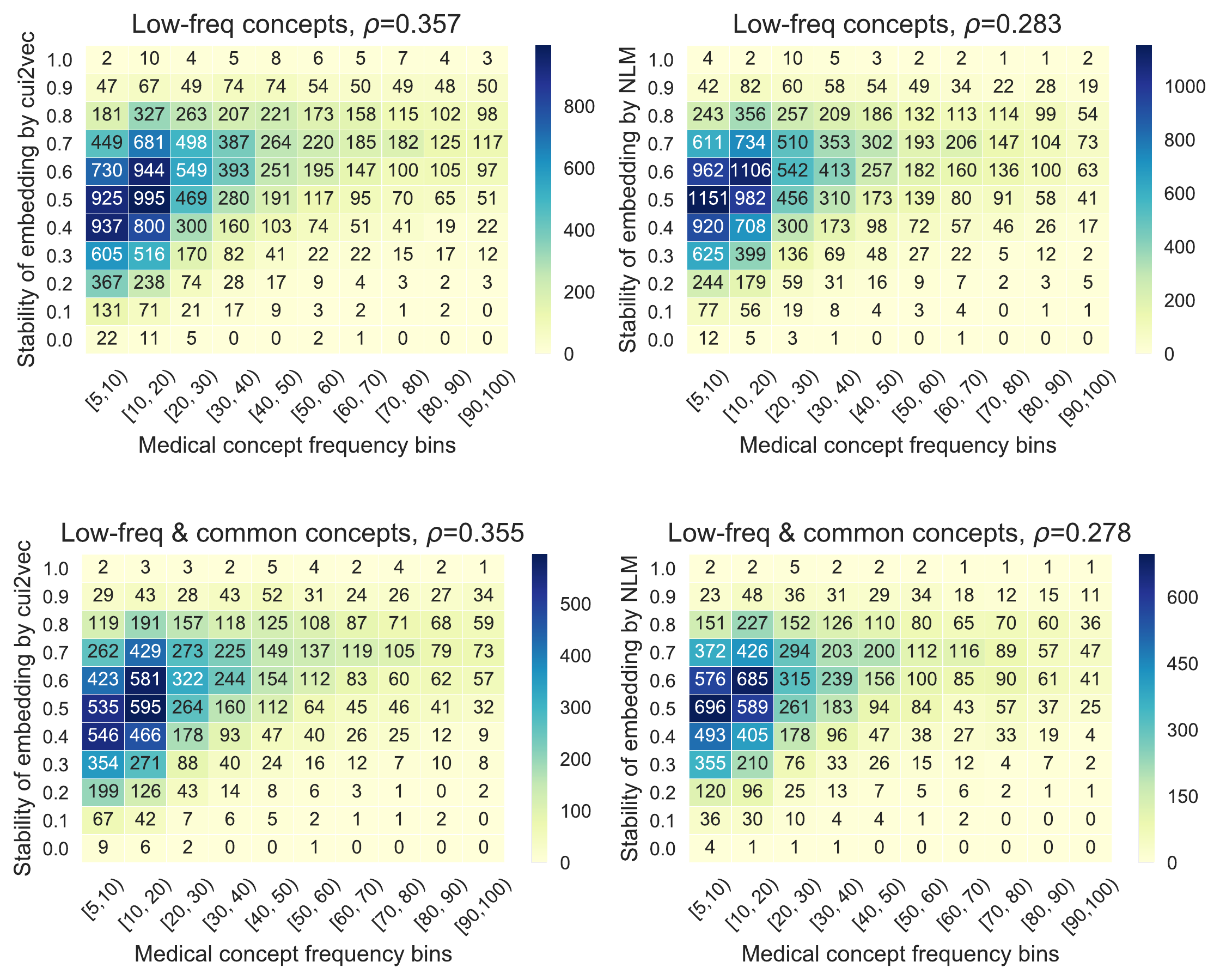}
	\caption{Distributions of low-frequency concepts (in the first row)  and commonly used concepts among low-frequency concepts (in the second row) over frequency and the stability of embeddings.}
	\label{fig:heatmaps_low_con}
\end{figure}

We take a close look at low-frequency concepts. We zoom into the first two columns in Figure~\ref{fig:heatmaps_all_con}, and present the distributions of low-frequency concepts alone in Figure~\ref{fig:heatmaps_low_con}.
We also provide the distributions of commonly used concepts among low-frequency concepts in Figure~\ref{fig:heatmaps_low_con}. 
In medical domain, concepts of diseases, symptoms, treatments, and diagnostic tests are commonly looked up by medical professionals~\cite{CIKM14_Limsopa_four_major_concepts}. 
Among all concepts (Figure~\ref{fig:heatmaps_all_con}), a large portion of concepts has a frequency lower than 100, and they are also important concepts (Figure~\ref{fig:heatmaps_low_con}). 
In Figure~\ref{fig:heatmaps_low_con}, more than a half of concepts have the stability greater than 0.5. 
Some low-frequency concepts have the stability even greater than 0.8.

We compute Pearson correlation coefficient ($\rho$) between concept frequency and embedding stability of all concepts. The correlation coefficient is 0.189 and 0.200 in cui2vec and NLM, respectively. 
Frequency shows a weak linear correlation with the embedding stability, because low-frequency concepts also have high stability as high-frequency concepts do.

\paratitle{Summary.}
In this section, we examine the stability of medical concept embeddings with frequency. High-frequency concepts have high-stability embeddings, showing that a high frequency indeed helps obtain quality embeddings with sufficient training instances. More importantly, we observe that surprisingly low-frequency concepts also have high-stability embeddings, despite the smaller number of training instances. 
The high-stability embeddings of low-frequency concepts are not aligned with the common understanding of the impact of frequency.
This results demonstrate that there might be other factors influencing the stability of concept embeddings in addition to frequency.

\section{Proposed Measure}
\label{sec:Noisiness_Stability}

We propose a new factor, \textit{the noisiness of context words}, which influences the stability of concept embeddings. Our hypothesis is as follows: if a concept has noisy context words, its embedding will undergo inconsistent learning, leading to a low-quality embedding. Conversely, if a concept has coherent context words, its embedding will go through consistent training, and thus it results in a high-quality embedding. 
We note that the proposed factor focuses on the distribution of context words only without considering the total number of context words, which is dependent on frequency. In the following discussion, we use `concept' and `word' interchangeably. The idea is the same.

To measure the noisiness of context words, we use \textit{normalized entropy}. Normalized entropy quantifies a distribution of elements (\ie context words in our context). It negates the effect of the total number of context words in the estimate using normalization. Formally, for a given word $w$, normalized entropy $H(w)$  is computed as follows.
\begin{equation}\label{Eq:normalized_entropy}
H(w)=-\sum_{w_i \in C_w} \dfrac {P(w_i)\cdot \log {P(w_i)}}{\log {|C_w|}}
\end{equation}
In the above equation, $C_w$ denotes a set of context words that co-occur with word $w$ within the window size. $P(w_i)$ is the relative frequency of context word $w_i$ in the collection of all context words. If context words of a target word are evenly distributed (flat distribution), normalized entropy is high; if a few context words account for a large portion in the entire context words (skewed distribution), the value is low.

We evaluate the proposed factor by computing a linear correlation coefficient (Pearson correlation coefficient) with the stability of concept embeddings. Table~\ref{tab:correl_result} presents coefficient values calculated in three groups of medical concepts: all concepts, low-frequency (frequency $<100$) concepts , and commonly used concepts among low-frequency concepts. For all concepts, the noisiness of context words shows a moderate linear correlation with the stability of concept embeddings. In general, the correlation is stronger in cui2vec than in NLM.

\begin{table} 
	\centering
	\caption{Pearson correlation coefficient of the noisiness of context words with the stability of medical concepts. We present three groups of medical concepts: all concepts, concepts with low-frequency (<100) (Low-freq), and commonly used concepts in biomedical applications among low-frequency concepts (Low-freq \& Common). All correlation coefficients are statistically significant at $p<0.001$.}
	\label{tab:correl_result}
	\begin{tabular}{c|c|c|c}
		\toprule
		& All concepts& Low-freq & Low-freq \& Common \\
		\midrule
		cui2vec & -0.593 & -0.540 &-0.549\\
		\midrule
		NLM & -0.477 & -0.370 & -0.329\\
		\bottomrule
	\end{tabular}
\end{table}

As shown in Figures~\ref{fig:heatmaps_all_con} and~\ref{fig:heatmaps_low_con}, while high-frequency concepts tend to have high stability, low-frequency concepts show varied the embedding stabilities. To better understand the surprising stability of low-frequency concepts, we look into low-frequency concepts and conduct a focused evaluation on them.

In Table~\ref{tab:correl_result}, the proposed factor consistently shows a moderate linear correlation with the stability of low-frequency concepts. The correlation in cui2vec is even comparable to the correlation estimated for all concepts. In NLM, the correlation is slightly decreased,  compared to the correlation for all concepts. 
The similar correlations are observed for commonly used concepts as well. 
Lastly, the proposed factor shows much stronger correlations with the stability of embeddings, compared to frequency as shown in Figures~\ref{fig:heatmaps_all_con} and \ref{fig:heatmaps_low_con}.
All correlation coefficients are statistically significant at the $p$-value smaller than 0.001.

\paratitle{Summary.}
In this section, we propose a new factor, the noisiness of context words, which influences the stability of medical concept embeddings. We use normalized entropy to estimate the proposed factor. The evaluation result shows a clear linear correlation between the proposed factor and the stability of medical concept embeddings. The correlations are consistent for all concepts,  for low-frequency concepts, and for commonly used low-frequency concepts. This result supports the claim that when a concept has coherent context words, it is likely to result in high-quality embeddings, regardless of high or low frequency in a corpus. Moreover, the proposed factor provides an empirical reasoning of the surprising high stability of low-frequency concepts.

\section{Suggestions on Improving the Stability}
\label{sec:Ex_low_stab_concepts}
We have shown the negative linear correlation in the evaluation. 
It indicates that decreasing the noisiness of context words improves the stability of medical concept embeddings. 
In this section, we make suggestions on ways to adjust the noisiness of context words, while maintaining the semantic meanings of concepts implied in a training corpus.

The first approach is to utilize hierarchical relations of medical concepts from a knowledge base. All medical concepts are linked to UMLS, and UMLS provides hierarchical relations among medical concepts. Hierarchical relations are a tree-like structure and connections indicate a parent-child (IsA) relation between concepts. In a parent-child relation, while the meaning of child concept is more specific than the meaning of parent concept, their meanings are greatly overlapped with each other. For example, \textsf{Knee replacement} (CUI: C0086511) is a child concept of  \textsf{Knee joint operation} (CUI: C0187769) in UMLS, and the meaning is knee joint operation for replacement. 
The noisiness of context words of low-stability concepts can be adjusted using the context words of parent and/or child concepts.

Another approach to decreasing the noisiness of context words is segregating a training corpus into smaller document clusters. Context words tend to be more consistent in topic-specific document corpus and  topic-specific sub-collection can be identified using document clustering techniques. Clustered documents provide information representing the  importance of context words, depending on which documents they appear. This information can be utilized to filter the context words. 

\section{Generalization}
\label{sec:Generalize_word}

We have evaluated the proposed factor with concept embeddings trained with medical textual data.
In this section, we conduct an experiment with embeddings trained on text from general domain.
We use two datasets: Reuters-21578 (Reuters) and Wikipedia abstract (Wiki).
Reuters consists of news articles under pre-defined topical categories. Wiki is much larger than Reuters and contains various topics.

In general domain, word embeddings are more prevalent than concept/entity embeddings. 
We train word embeddings without concept identification, and all trained word embeddings are subject to evaluations. 
Likewise, the skip-gram model with negative sampling is used. 
After the training, every word that has a frequency greater than minimum frequency threshold (set to 5) in a dataset has a word embedding.

Pearson correlation coefficients  on Reuters and Wiki are $-0.412$ and $-0.495$, respectively. 
The proposed factor shows the moderate correlations with the stability of word embeddings in both datasets.
The correlation values are statistically significant by $p<0.001$.
This result demonstrates that the linear relation of the proposed factor with the stability is not limited to medical-domain text, and it extends to general-domain text.

\section{Conclusion}
\label{sec:Conclusion}
We analyze the stability of medical concept embeddings with respect to frequency. The analysis shows that low-frequency concepts can achieve high stability even though there is the limited number of training instances for them. Motivated by the surprising high stability of low-frequency concepts, we propose a new factor, the noisiness of context words, influencing the stability of medical concept embeddings. We measure the noisiness of context words using normalized entropy.  The evaluation result shows a clear correlation between the proposed factor and the stability of medical concept embeddings. The result is consistent for all concepts, low-frequency concepts, and commonly used low-frequency concepts. This work helps understand how low-frequency concepts can result in high-stability embeddings like high-frequency concepts do.

\bibliographystyle{ACM-Reference-Format}


\section{Appendix}

In the skip-gram model, different hyperparameters affect the quality of word embeddings. We  evaluate the proposed factor on multiple sets of word embeddings that are trained with different hyperparameter values.
We test five hyperparameters: window size (W), epoch (E), number of negative samples in negative sampling (N),  smoothing parameter for negative sample distribution in negative sampling (M), and subsampling rate (S). 
Default values for W, E, N, M, and S are 7, 50, 5, 0.75, and 0.001, respectively. 
For each hyperparameter, two additional values are tested, while the rest hyperparameters are the default values.
Likewise, we train medical concept embeddings using the two approaches, cui2vec and NLM.

Table~\ref{tab:appendix_stability_hypers} presents the average and the standard deviation of the stability. 
Marginal changes are observed in both cui2vec and NLM when different hyperparameter values are used.

\begin{table}
	\centering
	\caption{Average (standard deviation) of the stability of medical concept embeddings when different hyperparameter values are set in the training of word embeddings}
	\label{tab:appendix_stability_hypers}
	\begin{tabular}{c|c|c|c}
		\toprule
		\textbf{Hyper-}& \multirow{2}{*}{\textbf{W:E:N:M:S}}& \multicolumn{2}{c}{\textbf{Avg (stdev) of stability}}\\
		\cline{3-4}
		\textbf{parameter}& & \textbf{cui2vec} & \textbf{NLM}\\
		\midrule
		\textit{Default}& \textit{7:50:5:0.75:0.001} &\textit{0.595 (0.208)} & \textit{0.627 (0.181)}\\
		\hline
		Window& 5 &0.595 (0.205) &0.636 (0.177) \\
		\cline{2-4}
		size (W)& 10 & 0.589 (0.210)& 0.611 (0.187)\\
		\midrule
		\multirow{2}{*}{Epoch (E)}& 30 & 0.564 (0.226)&0.622 (0.189)\\
		\cline{2-4}
		& 100 & 0.619 (0.197)&0.622 (0.182)\\
		\midrule
		Number of & 10 & 0.618 (0.203)&0.656 (0.177)\\
		\cline{2-4}
		NS (N)& 15 & 0.625 (0.202)&0.669 (0.175)\\
		\midrule
		Smoothing & 0.0 & 0.579 (0.229)&0.649 (0.179)\\
		\cline{2-4}
		 (M)& 1.0 & 0.568 (0.209)&0.611 (0.185)\\
		\midrule
		Subsampling & 0.01 & 0.604 (0.206)&0.632 (0.181)\\
		\cline{2-4}
		 rate (S)& 0.0001 & 0.579 (0.208)&0.609 (0.18)\\
		\bottomrule
	\end{tabular}
\end{table}

Table~\ref{tab:appendix_correl_hypers} shows Pearson correlation coefficient of the proposed factor with the stability when embeddings are trained with different hyperparameter values.
The correlation coefficients present small changes. 
This result makes sense that the stabilities are barely changed by the different hyperparameter values (Table~\ref{tab:appendix_stability_hypers})

Across the different hyperparameter values, the proposed factor shows a moderate linear correlation with the stability of medical concept embeddings.
It is worth noting the correlations with the window size (W). The different values of W directly changes the noisiness of context words as the window size determines context words.
When different window sizes are used in training of word embeddings, the proposed factor consistently shows the moderate correlations.

\begin{table}
	\centering
	\caption{Pearson correlation coefficients of the noisiness of context words with the stability of medical concpet embeddings. Embeddings are trained by different hyperparameter values. The strongest correlation in cui2vec and NLM is indicated as *. All correlation values are statistically significant at $p$ < 0.001.}
	\label{tab:appendix_correl_hypers}
	\begin{tabular}{c|c|c|c}
		\toprule
		\textbf{Hyper-}& \multirow{2}{*}{\textbf{W:E:N:M:S}}& \multicolumn{2}{c}{\textbf{Pearson coefficient}}\\
		\cline{3-4}
		\textbf{parameter}& & \textbf{cui2vec} & \textbf{NLM}\\
		\midrule
		\textit{Default}& \textit{7:50:5:0.75:0.001} &\textit{-0.593} &\textit{-0.477}\\
		\hline
		Window& 5 &-0.577 &-0.484  \\
		\cline{2-4}
		size (W)& 10 & -0.605&-0.470\\
		\midrule
		\multirow{2}{*}{Epoch (E)}& 30 & -0.618*& -0.478\\
		\cline{2-4}
		& 100 & -0.584& -0.501\\
		\midrule
	    Number of& 10 & -0.564&-0.471\\
		\cline{2-4}
		NS (N) & 15 & -0.556&-0.469\\
		\midrule
		Smoothing & 0.0 & -0.519& -0.335\\
		\cline{2-4}
		(M)& 1.0 & -0.601& -0.506*\\
		\midrule
		Subsampling & 0.01 &  -0.590&-0.484\\
		\cline{2-4}
		rate (S)& 0.0001 &  -0.578& -0.454\\
		\bottomrule
	\end{tabular}
\end{table}

\end{document}